\documentclass[10pt,twocolumn,letterpaper]{article}

\usepackage[pagenumbers]{cvpr} 

\usepackage{graphicx}
\usepackage{amsmath}
\usepackage{amssymb}
\usepackage{booktabs}
\usepackage{tablefootnote}

\usepackage[pagebackref,breaklinks,colorlinks]{hyperref}

\usepackage[capitalize]{cleveref}
\crefname{section}{Sec.}{Secs.}
\Crefname{section}{Section}{Sections}
\Crefname{table}{Table}{Tables}
\crefname{table}{Tab.}{Tabs.}

\def\cvprPaperID{5320} 
\def\confName{CVPR}
\def\confYear{2023}

\begin{document}

\title{Efficient Joint Detection and Multiple Object Tracking with Spatially Aware Transformer}

\author{Siddharth Sagar Nijhawan\\
Columbia University\\
{\tt\small sn2951@columbia.edu}
\and
Leo Hoshikawa\\
Sony Group Corporation\\
{\tt\small Leo.Hoshikawa@sony.com}
 \and
Atsushi Irie\\
Sony Group Corporation\\
{\tt\small Atsushi.Irie@sony.com}
\and
Masakazu Yoshimura\\
Sony Group Corporation\\
{\tt\small Masakazu.Yoshimura@sony.com}
\and
Junji Otsuka\\
Sony Group Corporation\\
{\tt\small Junji.Otsuka@sony.com}
\and
Takeshi Ohashi\\
Sony Group Corporation\\
{\tt\small Takeshi.a.Ohashi@sony.com}
}
\maketitle

\begin{abstract}
   We propose a light-weight and highly efficient Joint Detection and Tracking pipeline for the task of Multi-Object Tracking using a fully-transformer architecture. It is a modified version of TransTrack, which overcomes the computational bottleneck associated with its design, and at the same time, achieves state-of-the-art MOTA score of 73.20\%. The model design is driven by a transformer based backbone instead of CNN, which is highly scalable with the input resolution. We also propose a drop-in replacement for Feed Forward Network of transformer encoder layer, by using Butterfly Transform Operation to perform channel fusion and depth-wise convolution to learn spatial context within the feature maps, otherwise missing within the attention maps of the transformer. As a result of our modifications, we reduce the overall model size of TransTrack by 58.73\% and the complexity by 78.72\%. Therefore, we expect our design to provide novel perspectives for architecture optimization in future research related to multi-object tracking.
\end{abstract}

\section{Introduction}
\label{sec:intro}

The actively studied task of Multi-Object Tracking (MOT) is defined as the process of automatically following all the identified objects of interests in a given stream of frames. It serves a variety of use-cases ranging from robotics, pose estimation to surveillance based applications like intrusion detection. Many of these tasks require the MOT algorithm to be deployed on embedded systems with a light-weight implementation. Recent works like TransTrack \cite{transtrack} and TrackFormer \cite{trackformer} achieve state-of-the-art MOT performance, however, struggle to solve the computational bottleneck involved in their deployment. Therefore, there is a huge scope of improving the existing MOT architectures to build efficient models having low deployment cost in terms of size and complexity while maintaining good tracking accuracy.

\begin{figure}
  \centering
  \includegraphics[width=0.9\linewidth]{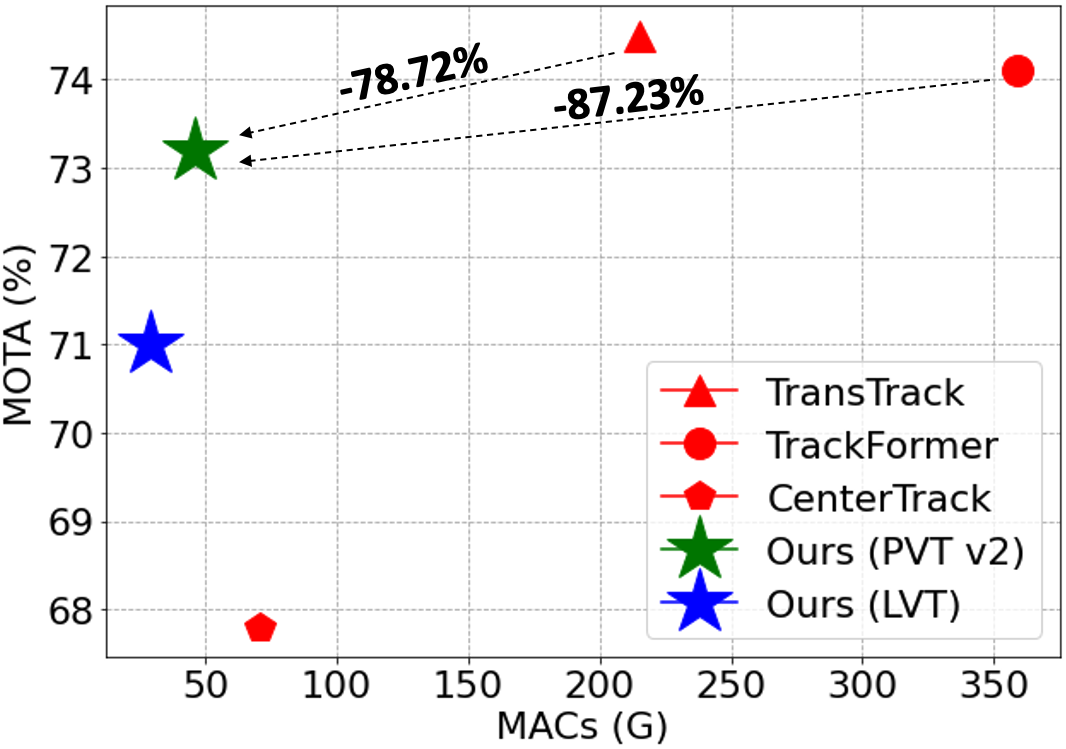}
   \caption{Comparison of our models with state-of-the-art across MACs vs. MOTA.}
   \label{fig:intro_res}
\end{figure}

MOT is computationally heavy because it involves multiple tasks: classification, object detection, and tracking. Since we can leverage good models for each of these tasks, Tracking by Detection (TBD) paradigm has been a predominant choice. It localizes all the objects of interests using a detector model and then associates each of these objects to tracks using information based re-identifiable features \cite{https://doi.org/10.48550/arxiv.1610.06136,Pang_2021_CVPR}. However, training a TBD system is highly complicated as it involves training each component separately \cite{gonzaleszuniga:hal-03541517}, making it difficult to be deployed on a system with limited computational resources. These challenges can be addressed with Joint Detection and Tracking (JDT), which performs detection and tracking simultaneously in a single stage \cite{8575355}. Unlike TBD, JDT trains each component jointly in a single end-to-end pipeline. JDT also achieves a significantly faster model speed compared to TBD, thus, making it as a suitable choice for designing a light-weight MOT architecture.

CNNs have been widely used to perform MOT by operating locally and greedily \cite{https://doi.org/10.48550/arxiv.1909.12605,Wojke_2018}. Such methods perform pairwise association between newly detected objects to existing set of confirmed tracks through distance metrics. Several techniques have recently emerged which leverage transformers \cite{NIPS2017_3f5ee243} and the query-key mechanism of JDT to perform MOT. TransTrack \cite{transtrack} introduces a set of learned object queries for performing detection of incoming objects using transformer encoder-decoder modules with CNN backbone and outperforms state-of-the-art architectures in MOT17 challenge \cite{https://doi.org/10.48550/arxiv.1603.00831}. A single training pipeline, fast post-processing, modular architecture, and high tracking performance favors TransTrack \cite{transtrack} to be a good baseline in the family of JDT architectures. 

Light-weight implementation of a model is dependent on two key characteristics: model size in terms of total trainable parameters and complexity, which is the total number of Multiplication Addition Accumulation Operations (MACs) required for a single forward pass. For the baseline of TransTrack \cite{transtrack}, an input of shape \((800 \times 1333)\) requires a total of 46.87M parameters and 215.23G MACs, which is a huge computational bottleneck. Therefore, we design a light-weight version of TransTrack \cite{transtrack} which overcomes the bottleneck with minimal to no impact on the tracking performance as shown in \cref{fig:intro_res}.

\begin{figure}[t]
  \centering
  \includegraphics[width=1.0\linewidth]{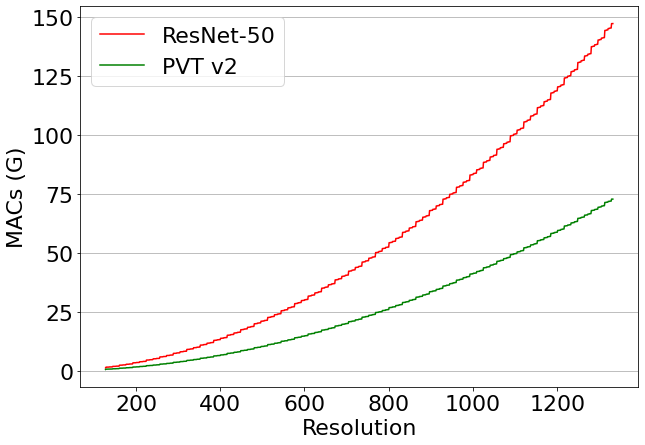}
   \caption{Input Resolution vs. MACs. PVT v2 \cite{pvtv2} requires almost half the MACs for a high input resolution of \((1333 \times 1333)\) and the slope gradually rises compared to ResNet-50 \cite{resnet}.}
   \label{fig:res_comparison}
\end{figure}

Some observations in TransTrack \cite{transtrack} can help us design a light-weight architecture with desirable characteristics. By performing layer-by-layer profiling in terms of parameters and MACs, we identified 2 key bottlenecks requiring 78\% total parameters with 92\% MACs: backbone and encoder. Most existing transformer based visual models, including TransTrack \cite{transtrack}, use the 'CNN + transformer' framework, where feature maps are extracted using a CNN and fed into transformer as an input. Since CNNs (ResNet-50 \cite{resnet} for TransTrack \cite{transtrack}) target local receptive fields due to the presence of convolution operation, they lack global context modeling. This can be only be solved through gradually expanding the receptive field by increasing the depth of convolution layer. However, doing so will drastically increase the model size and complexity. For dense prediction tasks like MOT, a larger input resolution leads to higher accuracy as it can generate rich representations of features related to small objects \cite{s21196358}. Vision transformers can generate global context through self-attention and also inherit the inductive bias of CNNs by spatially targeting local receptive fields within an input image. Architectures like Swin Transformer \cite{swin}, LVT \cite{lvt} and PVT \cite{pvt} serve as a general-purpose backbone for computer vision, having the flexibility to scale better in terms of computational complexity with respect to image size. LVT \cite{lvt} introduces convolution in self-attention operation to extract low-level features. PVT v2 \cite{trackformer}, an improved version of PVT \cite{transtrack}, reduces the computational complexity to linear and achieves significant improvements on fundamental vision tasks through a progressive shrinking pyramid. \cref{fig:res_comparison} validates the scalability of PVT v2 \cite{trackformer} (b1 configuration) in comparison to ResNet-50 when the input resolution is increased. Therefore, we propose PVT v2 \cite{trackformer} as a direct replacement for existing backbone in TransTrack \cite{transtrack} leading to a significant reduction in model size and complexity.

Encoder contains self-attention mechanism which enables it to find global dependencies among feature maps but it loses the spatial information contained within feature maps as they are flattened into patch embeddings. Inspired by this problem, we propose a new block based on depth-wise convolution similar to MobileNet \cite{https://doi.org/10.48550/arxiv.1704.04861}. Our block replaces the feed forward network (FFN) of the encoder with 3 sub-blocks: 2 channel fusion blocks using Butterfly Transform (BT) \cite{bft}, and a depth-wise convolution block. BT \cite{bft} is a drop-in replacement for channel fusion operation with logarithmic complexity rather than quadratic. Our proposed block solves 2 key problems. It enables the encoder to learn spatial information by converting patch embeddings back to two dimensional feature maps and passing them through depth-wise convolution to learn spatial context. By replacing feed forward network which is the key bottleneck in encoder due to dimensionality expansion by a factor of 8, our block reduces the complexity significantly by 86.76\%. To sum up, the contributions of this paper are as follows:

\begin{figure*}[t]
\centering
\begin{subfigure}[b]{0.9\textwidth}
   \includegraphics[width=1\linewidth]{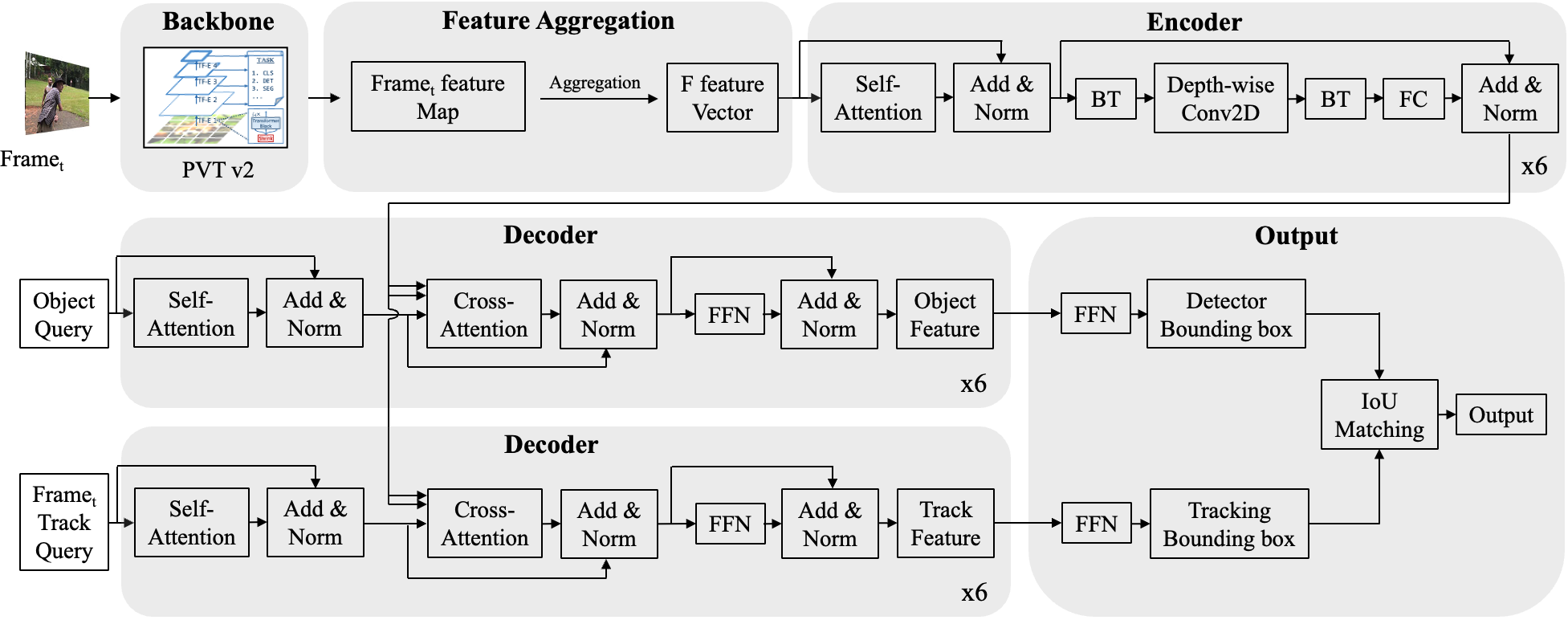}
   \caption{}
   \label{} 
\end{subfigure}

\begin{subfigure}[b]{0.55\textwidth}
   \includegraphics[width=1\linewidth]{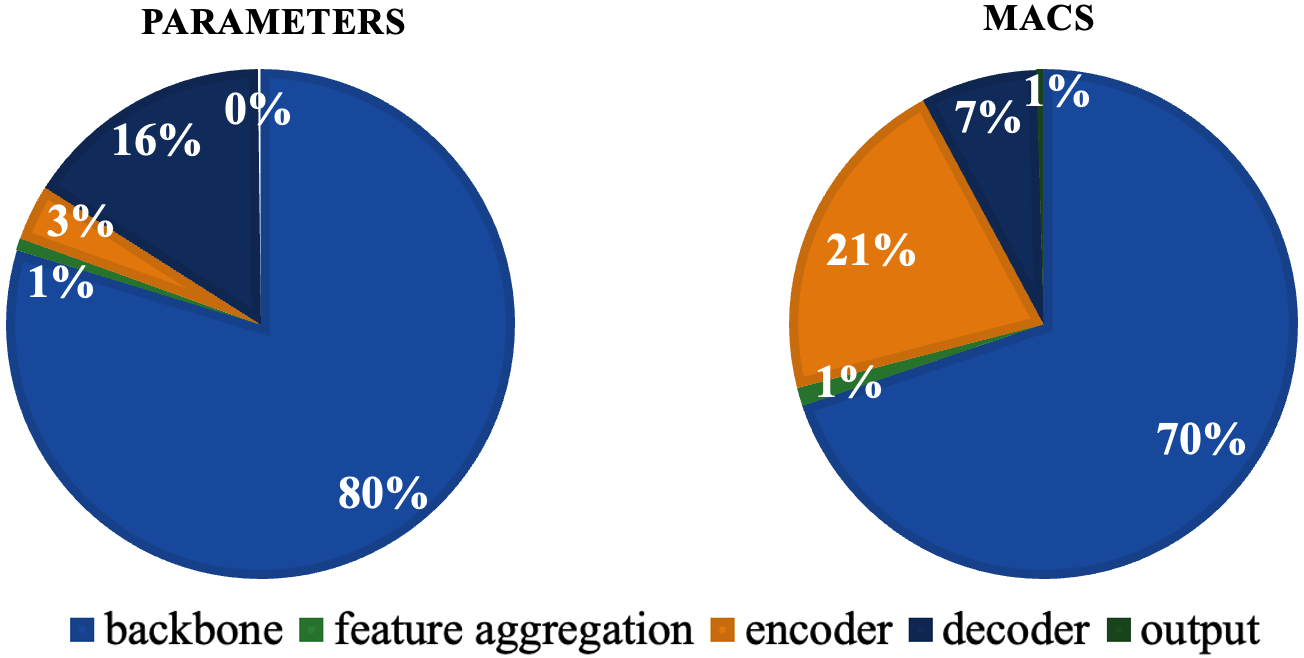}
   \caption{}
   \label{}
\end{subfigure}
\caption{(a) Architecture of proposed Multi-object Tracker and (b) Layer-by-layer profiling of our model with PVT v2 \cite{pvtv2} backbone.}
\label{fig:architecture}
\end{figure*}

\begin{itemize}
	\item We performed layer-by-layer profiling of TransTrack \cite{transtrack} to identify key bottlenecks and target these modules to reduce the total number of parameters and overall complexity of the model significantly.
	
	\item Through ablation studies, we showcased that CNN backbone is not computationally efficient for high input resolution and a transformer-based backbone scales better in generating feature maps for dense prediction tasks.
	
	\item We solved the computational bottleneck associated with encoder block by proposing a block which performs channel fusion operation with logarithmic space/time complexity through BT operation \cite{bft} and at the same time, learns the spatial context within the feature maps. As a result, proposed block requires 74.3\% less MACs compared to a standard transformer encoder. 

	\item By applying the above mentioned modifications, our architecture contains 58.73\% less parameters and requires 78.72\% less MACs in comparison to TransTrack \cite{transtrack}, while achieving state-of-the-art MOT score of 73.20\%. To our knowledge, this is the first fully-transformer based JDT pipeline. In conclusion, our model is an extremely light-weight version of TransTrack \cite{transtrack} with small deployment costs.
\end{itemize}

\section{Related Work}
\label{sec:relatedwork}

\textbf{Vision Transformers.} Numerous techniques have used transformers for image classification \cite{vit} and other dense prediction tasks with efficient usage of resources. Self-Attention is performed using a shifting windows approach in Swin Transformer \cite{swin}, which leverages local windows instead of global attention to efficiently compute attention within images. Similarly, LocalViT \cite{localvit} integrates depth-wise convolution into the encoder architecture to enable local understanding in the image.  PVT \cite{pvt} and PVT v2 \cite{pvtv2} form a feature pyramid network which performs down-sampling across four stages of feature maps. Another light-weight architecture, MobileViT \cite{mobilevit} treats transformers as convolutions to learn global representations with mobile-level resource requirements.

\textbf{Tracking.} Appearance features are extracted through deep-learning based algorithms and passed to detector phase in the TBD methodology of POI \cite{https://doi.org/10.48550/arxiv.1610.06136}. However, occlusion due to crowded scenes is the biggest challenge for such frameworks. Contrary to SDT, JDT methods like \cite{8575355} use Single Shot MultiBox (SSD) object detector \cite{Liu_2016} on a single frame and feeds the output to recurrent neural network for integrating detections into tracks. CenterTrack \cite{https://doi.org/10.48550/arxiv.2004.01177} performs tracking-conditioned detection to predict object offsets. GTR \cite{gtr} designs a global tracking transformer architecture to produce a global set of trajectories for all objects. FairMOT \cite{Zhang_2021} trains an instance classification branch together with the detector and performs association via pairwise ReID features. It achieves good performance, however, requires a very heavy post-processing stage. As a JDT framework based on transformer, TransTrack \cite{transtrack} leverages the information of previous frame to generate object association in the current frame and performs set prediction for detection at the same time. It contains two sets of keys: object and track queries. Object queries are learned similar to transformer-based detectors \cite{https://doi.org/10.48550/arxiv.2005.12872} with an encoder decoder pair. Track queries are learned from the object features of previous frame generated through the CNN backbone of ResNet-50 \cite{resnet}. TransTrack \cite{transtrack} uses a simple technique of IoU matching to generate the final set of objects with bounding boxes and IDs. Similarly, the proposed framework performs JDT by treating tracked features from previous frame as the query for current frame and matches them with the object detections as the key.

\textbf{Architecture Optimization.} The amount of computation involved in a model can be reduced significantly by designing compact network architectures. MobileNet \cite{https://doi.org/10.48550/arxiv.1704.04861} reduces the complexity of computing convolution operation in CNNs through depth-wise separable convolution. ShuffleNet \cite{https://doi.org/10.48550/arxiv.1707.01083} introduces channel shuffle operation to reduce the computation of \(1 \times 1\) convolution, resulting in highly efficient networks outperforming architectures like ResNet \cite{resnet}. Several techniques have been proposed to optimize the computational overhead associated with transformers. Linformer \cite{https://doi.org/10.48550/arxiv.2006.04768} approximates the self-attention mechanism by a low-rank matrix reducing the complexity from quadratic to linear. PVT v2 \cite{pvtv2} improves the efficiency by using a pyramid structure with linear attention module. \cite{https://doi.org/10.48550/arxiv.2103.11816} proposes a locally-enhanced FFN to promote correlation among tokens and spatial dimensions using depth-wise convolutions.


\section{Proposed Methodology}
\label{sec:proposedmethod}

\begin{figure}[t]
  \centering
   \includegraphics[width=1.0\linewidth]{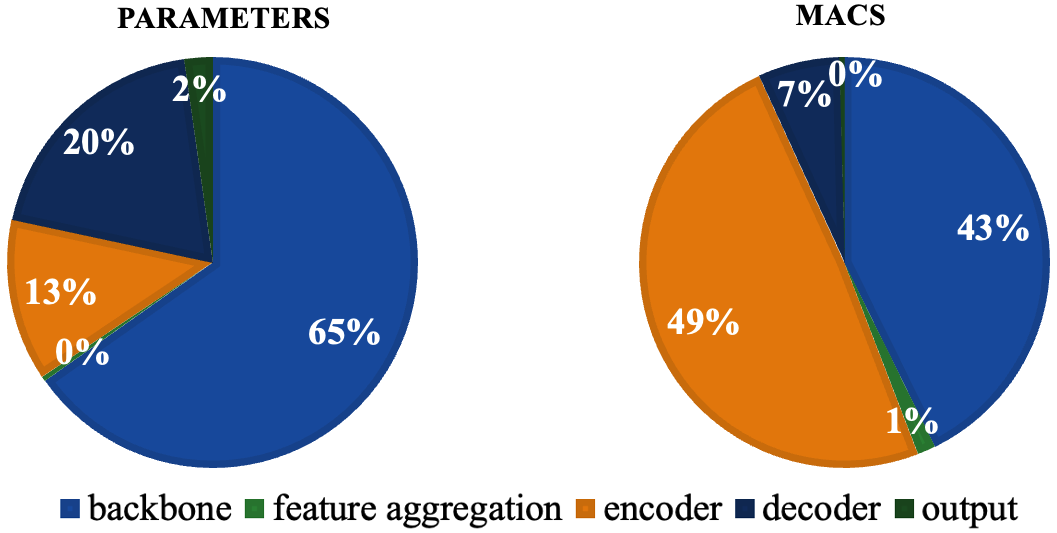}
   \caption{Layer-by-layer profiling of TransTrack \cite{transtrack} architecture in terms of Trainable Parameters and MACs required for a single forward pass.}
   \label{fig:transtrack_profiling}
\end{figure}

The proposed architecture is described in \cref{fig:architecture} which is based on TransTrack \cite{transtrack} with several improvements to reduce the computational complexity and model size. On a high-level, we generate a set of bounding boxes per objects of interests in each frame. These boxes are predicted through object and track queries learned during the training process. Following the JDT paradigm of TransTrack \cite{transtrack}, these two sets of bounding boxes are called detection and tracking bounding boxes. At the final stage, these are associated with each other through Hungarian algorithm \cite{https://doi.org/10.1002/nav.3800020109} to generate final set of bounding boxes.

\subsection{Problem Formulation}
To identify the areas of improvement in TransTrack \cite{transtrack}, we passed a random tensor of shape \((3 \times 800 \times 1333)\) and profiled the number of trainable parameters per layer and total MACs required for a single forward pass. \cref{fig:transtrack_profiling} shows the percentage distribution of these metrics across 5 key layers: backbone, feature aggregation, encoder, decoder, and output. Backbone consists of ResNet-50 \cite{resnet} architecture which generates 4 feature maps of the current frame and 4 of previous frame aggregated to a single vector using feature aggregation layer. These feature maps are passed on to encoder, followed by a set of decoders to perform detection and tracking. Finally, output is generated through output layer of IoU matching. We conclude that model size is majorly dependent on ResNet-50 \cite{resnet} backbone with 23.23M parameters consisting of 65\% of the total size. In terms of computational complexity, the key bottleneck is the encoder which required 49\% of total MACs. Therefore, our architecture optimizes two key components of TransTrack \cite{transtrack} to make it light-weight; backbone and encoder. Next section describes each of the components of our proposed architecture in detail.

\begin{figure*}[t]
  \centering
  \includegraphics[width=1.0\linewidth]{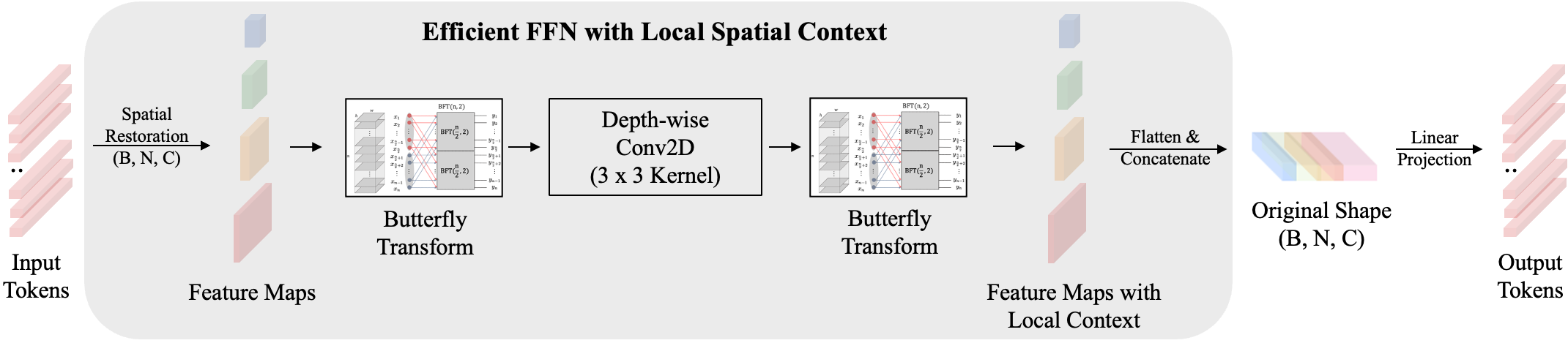}
  \caption{Proposed replacement for encoder's FFN.}
  \label{fig:depthwise}
\end{figure*}

\subsection{Architecture Components}

This section describes each component for extracting spatial feature maps, performing object detection, generating tracking boxes, and finally associating them to form final set of output boxes.

\textbf{Backbone.} The first step in JDT is to pass the current frame to the backbone of the network which functions as the feature map extractor. PVT v2 \cite{pvtv2} backbone of our model incorporates a progressive shrinking pyramid by uniformly reducing spatial dimensions at each stage which reduces the length of input embedding sequence as we go deeper into the network, therefore, reducing the computational cost associated with it. It can also be easily plugged into several existing dense prediction pipelines without any major modifications to the structure \cite{pvt}. PVT v2 \cite{pvtv2} uses patch embedding instead of pixel level features and requires lower number of MACs \cite{pvt}. Therefore, PVT v2 \cite{pvtv2} is a suitable replacement for ResNet-50 \cite{resnet} in generating feature maps with low computational cost. To optimize the computation of feature map extraction, we temporarily save the extracted maps of current frame to be used for the next frame. 

\textbf{Encoder.}\ We propose a modification in transformer encoder structure of TransTrack \cite{transtrack}. The FFN of encoder is computationally very expensive because it expands the dimensions by a factor of \(8\). Also, the attention mechanism of encoder lacks the understanding of spatial context among image patch tokens. Therefore, we replace the first fully connected (FC) layer with a structure shown in \cref{fig:depthwise} containing three sub-blocks: two BT operations and a single \(3 \times 3\) depth-wise convolution block. After generating a set of tokens from previous Multi-head Self Attention (MHSA) block, the patch tokens are restored back to spatial dimensions with shapes relative to four feature maps generated by the backbone. Then we perform depth-wise separable convolution \cite{https://doi.org/10.48550/arxiv.1610.02357}, however, performing channel fusion with BT blocks rather than \(1 \times 1\) point-wise convolution. We use a \(3 \times 3\) kernel for performing depth-wise convolution, enhancing the correlation among neighboring 8 pixels for spatial representation. After fusing spatial context, we flatten these tokens into original sequence of shape \(B \times N \times C\), where B is the batch size, N is the sequence length, and C is the total number of channels. Finally, we add a linear projection layer without expanding the channel dimensions. The function of this projection layer is to process the output of current layer for the next attention layer by parameterizing these modules. 

We use BT \cite{bft} operation to replace channel fusion in depth-wise separable convolution block to reduce the computational complexity of \(1 \times 1\) point-wise convolution which is considerably higher than spatial fusion operation of \(3 \times 3\) depth-wise convolution \cite{https://doi.org/10.48550/arxiv.1704.04861}. Our block treats channel fusion as a matrix-vector product problem defined in \cref{eq:matrix_product}. 

\begin{equation} \label{eq:matrix_product}
\hat{Y} = \hat{W} \hat{X}
\end{equation}

Here, \(\hat{Y}\) is the matrix representation of output tensor \(Y\). \(\hat{X}\) and \(\hat{W}\) are 2-D matrices generated by reshaping input tensor \(X\) and weight tensor \(W\) respectively. To perform this matrix multiplication and fuse information across all channels, we generate several sequential layers, analogous to a fast-fourier transform operation. We generate partitions of input channels and output channels of the same layer and connect each of these with parallel edges. Therefore, the output contains information from all the channels. This process is recursively repeated for each output point in the remaining layers. The operation reduces the space and time complexity of \(1 \times 1\) point-wise convolution from \(O(N^2 WH)\) to \(O((NlogN)WH)\), where N is the number of input and output channels. Therefore, our block combines the advantages of both CNN and transformer into a single encoder module by extracting local information and keeping the self-attention portion unchanged. It also reduces the number of trainable parameters in each encoder block from 4.54M in TransTrack \cite{transtrack} to 602.11K. To perform a single forward pass, the encoder in TransTrack \cite{transtrack} required 100.49G MACs, whereas, our block only requires 13.30G MACs which is a huge \(86.76\%\) reduction.

\textbf{Detection.}\ Trajectory Query Decoder performs object detection to generate detection bounding boxes based on the architecture proposed in Deformable DETR \cite{https://doi.org/10.48550/arxiv.2010.04159}. We feed object query to the decoder which is a learnable parameter trained along with all the other parameters of the network. Decoder performs a look-up operation between the aggregated feature maps of the input image as the key and object queries as the query, and looks up objects of interests in the frame. The output of this stage is a set of detection bounding boxes.

\textbf{Tracking.}\ This stage takes the input as detected object in the previous frame and passes them as object features on to the next frame. These object features are termed as 'track queries' and the decoder has the same architecture as Detection stage, however, with different inputs. Therefore, decoder takes both spatial appearance based features along with location based features of objects in the previous frame, and looks up the coordinates of these objects in the current frame to output tracking bounding boxes.

\textbf{Association.}\ The final stage performs pairwise mapping of the boxes generated from both decoder stages using IoU mapping and generates the output bounding boxes. The boxes which have no match are initialized as new object tracks.

\textbf{Training Loss.}\ Similar to TransTrack \cite{transtrack}, we train the two sets of decoders together using the same training loss as they produce bounding boxes in the same frame. Therefore, detection as well as tracking boxes are supervised using set prediction loss. This facilitates bipartite matching of ground truth objects and the predicted object trajectories in the same way as implemented in TransTrack \cite{transtrack}. The cost \(L\) for the bipartite matching problem is defined in \cref{eq:loss}.

\begin{equation} \label{eq:loss}
L = \lambda\textsubscript{class} L\textsubscript{class} + \lambda\textsubscript{iou} L\textsubscript{iou} + \lambda\textsubscript{giou} L\textsubscript{giou}
\end{equation}

\(L\textsubscript{class}\) is defined as the loss between ground truth class labels and predicted classifications. The generalized IoU loss \cite{https://doi.org/10.48550/arxiv.1902.09630} and bounding-box regression loss is defined by \(L\textsubscript{giou}\) and \(L\textsubscript{iou}\) respectively. This is computed between the center coordinates and height / width of predicted bounding boxes and ground truth boxes after normalization. The remaining terms are simply the coefficients assigned to each term of the loss. Since we perform matching on pairs, the training loss is same as association loss for each of these pairs. To compute the final loss value, we simply add all the individual pair losses and normalize it by dividing through the total number of objects per batch.

\textbf{Inference.}\ The first step during inference is to detect objects in the frame, provided that we have the feature maps of the current and previous frames. We compute the object trajectories by propagating these object features through next frame and performing box association. This process is repeated for the entire video sequence. Similar to TransTrack \cite{transtrack}, occlusions are handled by using track rebirth during inference. We keep a track of how many frames does a tracking box remain unmatched for a total of 32 consecutive frames. In this way, unmatched tracking boxes are matched to detection boxes and assigned their unique IDs.

\begin{figure}[t]
  \centering
   \includegraphics[width=1.0\linewidth]{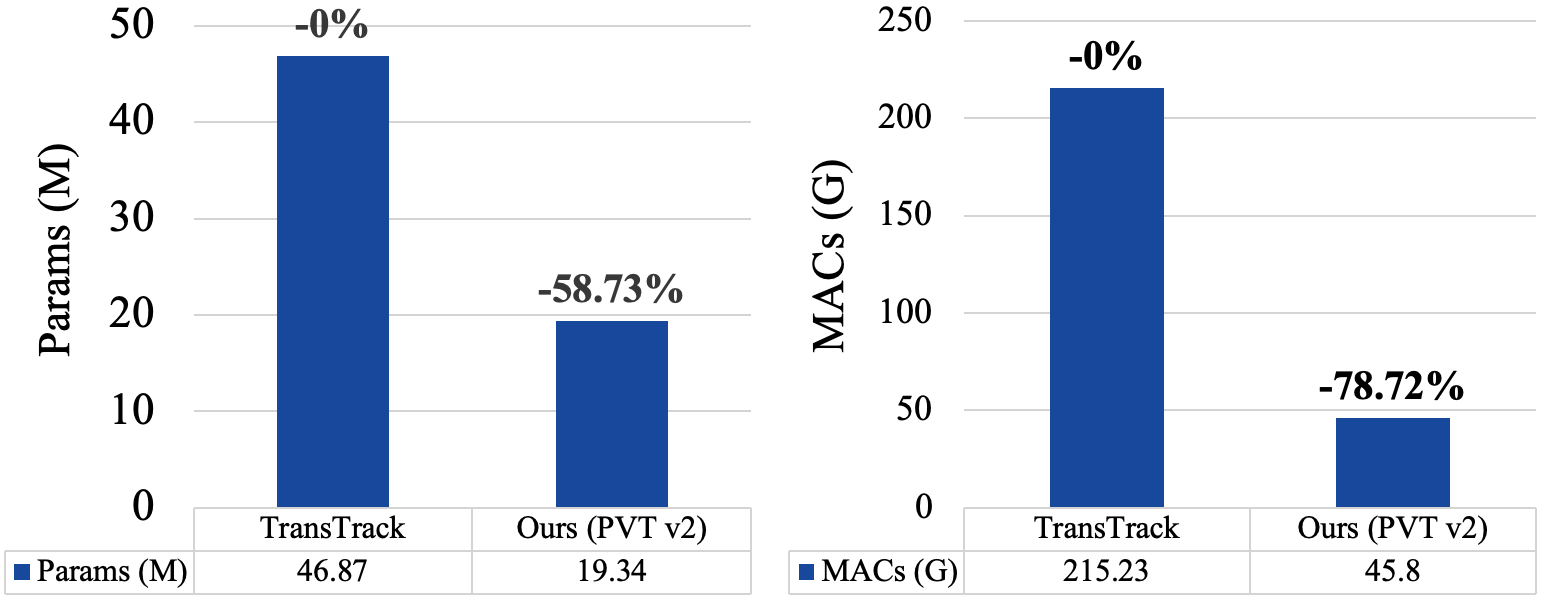}
   \caption{Comparison of Trainable Parameters and MACs required for a single forward pass between TransTrack \cite{transtrack} and proposed method.}
   \label{fig:profiling_comparison}
\end{figure}

\subsection{Comparison with TransTrack}

\cref{fig:profiling_comparison} indicates the overall profiling of TransTrack's \cite{transtrack} JDT architecture v.s. the proposed design with various modifications. It is evident that proposed model is highly efficient in terms of computational complexity, requiring 78.72\% less MACs compared to TransTrack. Also, the proposed model is very light-weight, containing 58.73\% less trainable parameters. We investigate the impact of proposed modifications on tracking performance in the next section.

\section{Experiments}
\label{sec:experiments}

\subsection{Training Setup}

\textbf{Datasets.}\ Training pipeline utilizes 2 datasets. First is CrowdHuman \cite{shao2018crowdhuman} which is a benchmark dataset to evaluate detectors in crowd scenarios. It has 15,000 training, 4,370 validation and 5,000 testing image stills without any temporal context. Due to a presence of 470,000 total human instances, it is a suitable choice of training human detectors. Second dataset is MOT17 \cite{https://doi.org/10.48550/arxiv.1603.00831}, which is suited for MOT along with other applications like re-identification containing 7 different indoor and outdoor scenes of public places with pedestrians as objects of interest. It provides detections of objects with 3 detectors containing around 15,948 training and 17,757 testing frames with varying input resolutions. We also use a subset of MOT17 data called ’MOT17 half-half' split which uses a single detector sequences having first half of video sequence as training and second half as validation data.

\textbf{Training Pipeline.}\ The training pipeline is divided into 4 phases similar to TransTrack \cite{transtrack}. For Phase 0, we pre-train the backbone on ImageNet dataset \cite{5206848} by randomly initializing the weights. Phase I involves pre-training the encoder and the detector on CrowdHuman \cite{shao2018crowdhuman} dataset with pre-trained backbone weights. Phase II trains the end-to-end MOT model on a mix of CrowdHuman \cite{shao2018crowdhuman} and MOT17 half-half split. We perform hyper parameter tuning using MOT17's half-half split set, and in Phase IV, we fine-tune the entire model on a mix of CrowdHuman \cite{shao2018crowdhuman} and MOT17's full train set. Finally, the evaluation is performed using MOT17's private test set.

\begin{table*}[t]
\centering
\caption{Quantitative Evaluation on MOT17 private test set.}
\resizebox{0.9\textwidth}{!}
{
\begin{tabular}{lcccccc} 
 \hline
 Model & MOTA$\uparrow$ & FP$\downarrow$ & FN$\downarrow$ & IDS$\downarrow$ & Params$\downarrow$ & MACs$\downarrow$ \\
 \hline
 TransTrack \cite{transtrack} & 74.50 & 28,323 & 112,137 & 3,663 & 46.87M & 215.23G \\
 TrackFormer \cite{trackformer} & 74.10 & 34,602 & 108,777 & 2,829 & 44.01M & 358.85G \\
 GTR \cite{gtr} & \textbf{75.30} & 36,231 & \textbf{93,150} & \textbf{2,346} & 43.80M & -\tablefootnote{Due to incompatibilities in the implementation, we could not compute the MACs for GTR.} \\
 CenterTrack \cite{https://doi.org/10.48550/arxiv.2004.01177}  & 67.80 & \textbf{18,489} & 160,332 & 3,039 & 19.32M & 70.88G \\
 FairMOT\cite{Zhang_2021}\tablefootnote{FairMOT does not include post-processing computation}  & 73.70 & 27,507 & 117,477 & 3,303 & 19.71M & 84.98G \\
 Ours (PVT v2 \cite{pvtv2}) & 73.20 & 28,341 & 118,689 & 4,218 & 19.34M & 45.80G \\
 Ours (LVT \cite{lvt}) & 71.00 & 32,730 & 125,274 & 5,757 & \textbf{6.07M} & \textbf{28.82G} \\
 \hline
 \label{table:mota_table}
\end{tabular}
}
\end{table*}

\subsection{Implementation details} \label{implementation_details}
We initialize the backbone with weights obtained through pretraining on ImageNet \cite{5206848}, shared publicly by PVT v2 \cite{pvtv2}. For the remaining portion of the architecture including detector and tracker, we use Xavier-init \cite{pmlr-v9-glorot10a} to initialize the weights, and fine-tune the model using AdamW \cite{https://doi.org/10.48550/arxiv.1610.02357} optimizer. The initial learning rate is set to 2e-4. Similar to TransTrack \cite{transtrack}, we apply various data augmentation techniques of random cropping, scaling, and re-sizing the inputs ranging from \((480 \times 800)\) to \((800 \times 1333)\) pixels. The learning rate scheduler is set to drop by a factor of 10 at 100th epoch, with a total of 150 epochs for training. Following the TransTrack \cite{transtrack} procedure, the end-to-end model is first pre-trained on CrowdHuman \cite{shao2018crowdhuman} and finally, fine-tuned on MOT17.

\subsection{Evaluation Metrics}
We evaluate our models quantitatively by computing MOTA score using \cref{eq:mota}.

\begin{equation} \label{eq:mota}
MOTA = 1 - \frac{ \sum_{t}^{} (FN\textsubscript{t} + FP\textsubscript{t} + IDS\textsubscript{t}) }{ \sum_{t}^{} GT\textsubscript{t} }
\end{equation}

For a frame t, \(GT\textsubscript{t}\), \(FN\textsubscript{t}\), \(FP\textsubscript{t}\), and \(IDS\textsubscript{t}\) denote the total number of ground truth objects, false negatives, false positives, and ID switches respectively. We also study the model properties, specifically, total number of trainable parameters and MACs required for a single forward pass. The evaluation is performed on MOT17 \cite{https://doi.org/10.48550/arxiv.1603.00831} private test set containing a total of 21 sequences with \(564,228\) boxes.

\subsection{MOT17 Benchmark}

Table \ref{table:mota_table} contains the evaluation results on MOT17 private test set for the proposed model along with several state-of-the-art MOT architectures. Our model with PVT v2 \cite{pvtv2} backbone achieves performance comparable to state-of-the-art methods with considerably lower model size and computational complexity. It has 58.73\% less trainable parameters and requires 78.72\% less MACs for a single forward pass in comparison to the baseline of TransTrack \cite{transtrack} with only 1.3\% drop in MOTA score. Similarly, it only has a 0.9\% difference in MOTA with TrackFormer \cite{trackformer} which requires 358.85G MACs whereas our model requires only 45.80G MACs. Compared to FairMOT, our proposal do not require heavy post-processing. We also test an extended version of our model which contains LVT \cite{lvt} backbone instead of PVT v2 \cite{pvtv2} and it further reduces the overall parameter size to 6.07M with 28.82G MACs. Both of our models outperform CNN based CenterTrack \cite{https://doi.org/10.48550/arxiv.2004.01177} with a relative improvement of 5.4\% and 3.2\% MOTA score for PVT v2 \cite{pvtv2} and LVT \cite{lvt} backbones respectively. We also achieve excellent FP and FN denoting that most of the objects within sequences were successfully detected. In terms of IDS, our models under-perform, however, it is a future direction of research to improvise the performance of our JDT architecture.

\section{Design choice experiments}

This section covers various ablation studies we performed on our key design choices. All the experiments are conducted on MOT17 half-half split with the setup described in \ref{implementation_details}. 

\begin{table}
\centering
\caption{Ablation study on architecture backbone.}
{
\begin{tabular}{lcccc} 
 \hline
 Backbone & MOTA$\uparrow$ & Params$\downarrow$ & MACs$\downarrow$ \\
 \hline
 ResNet-50 \cite{resnet} & \textbf{66.30} & 30.12M & 80.04G \\
 PVT v2 \cite{pvtv2} & 65.70 & 19.34M & 45.80G \\
 LVT \cite{lvt} & 64.60 & \textbf{6.07M} & \textbf{28.82G} \\
 \hline
 \label{table:ablation_backbone}
\end{tabular}
}
\end{table}

\begin{table}
\centering
\caption{Ablation study on number of encoders.}
{
\begin{tabular}{ccccc} 
 \hline
 Encoders & MOTA$\uparrow$ & Params$\downarrow$ & MACs$\downarrow$ \\
 \hline
 1 & 49.10 & 19.15M & 37.76G \\
 3 & 49.80 & 19.43M & 49.80G \\
 6 & 67.00 & 19.84M & 62.18G \\
 \hline
 \label{table:ablation_num_encoders}
\end{tabular}
}
\end{table}

\begin{table}
\centering
\caption{Ablation study on number of decoders.}
{
\begin{tabular}{ccccc} 
 \hline
 Decoders & MOTA$\uparrow$ & Params$\downarrow$ & MACs$\downarrow$ \\
 \hline
 1 & 55.10 & 15.53M & 42.82G \\
 3 & 64.45 & 17.25M & 44.01G \\
 6 & 67.00 & 19.84M & 62.18G \\
 \hline
 \label{table:ablation_num_decoders}
\end{tabular}
}
\end{table}

\subsection{Backbone} \label{ablation_backbone}

We ablate the effect of changing the backbone in our architecture with 3 design choices: ResNet-50 \cite{resnet}, PVT v2 \cite{pvtv2} and LVT \cite{lvt}. The quantitative results are shown in Table \ref{table:ablation_backbone}. LVT \cite{lvt} backbone is the most light-weight and efficient out of our backbone choices, with only 6.07M parameters and 28.82G MACs, however, achieves 64.60\% MOTA on MOT17 half-half test set. PVT v2 \cite{pvtv2} slightly performs better, however, consumes more than 50\% parameters and MACs. Choosing the most widely used backbone of ResNet-50 \cite{resnet} is highly inefficient, with 80.04G MACs and 30.12M parameters, proving our hypothesis that transformer backbones scale better with resolution as compared to CNN backbones.

\subsection{Number of encoders and decoders} \label{ablation_encoder_decoder}

Using the proposed architecture with PVT v2 \cite{pvtv2} backbone, we study the impact of number of encoders and decoders on MOTA, model size and efficiency in Table \ref{table:ablation_num_encoders} and Table \ref{table:ablation_num_decoders} respectively. Using a single encoder reduces the performance significantly, achieving only 49.10\% MOTA instead of 67.00\% for a total of 6 encoders. Similar trend is observed for decoders, where a single decoder achieves 55.10\% MOTA with a total of 42.82G MACs. To maximize the performance in terms of MOTA, we choose 6 encoders and 6 decoders as default for our experiments.

\begin{figure}[t]
  \centering
   \includegraphics[width=1.0\linewidth]{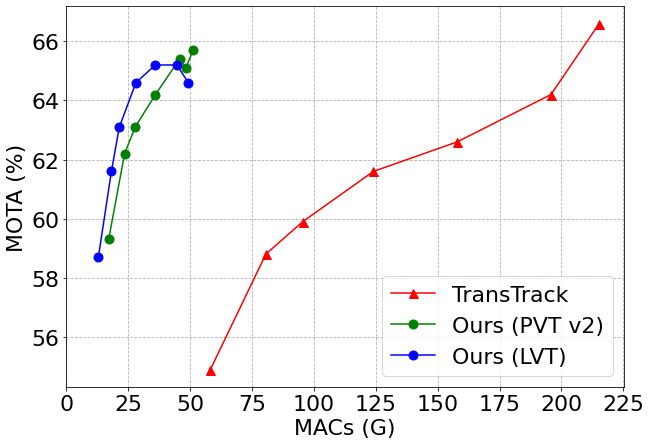}
   \caption{Ablation study on impact of input resolution on MOTA and MACs required for a single forward pass.}
   \label{fig:ablation_res}
\end{figure}

\subsection{Input Resolution} \label{ablation_res}

The computational complexity of the model is largely dependent on the input resolution. Also, larger the input resolution, the better the tracking performance \cite{s21196358}. To validate this, we ablate the dimensions of input across 7 choices,  \((400 \times 666)\), \((480 \times 799)\), \((520 \times 866)\), \((600 \times 999)\), \((680 \times 1133)\), \((760 \times 1266)\) and \((800 \times 1333)\). Dimension \((800 \times 1333)\) is the inference resolution proposed by TransTrack \cite{transtrack}. \cref{fig:ablation_res} shows the relationship between the tracking performance and MACs for chosen input resolutions. Both our models are highly efficient across all the choices of resolutions compared to TransTrack \cite{transtrack}. For an input resolution of \((800 \times 1333)\), our model with PVT v2 \cite{pvtv2} backbone requires 51.23G MACs and with LVT \cite{lvt} backbone requires 49.34G MACs. On the other hand, TransTrack \cite{transtrack} needs approximately 4 times more resources at this resolution, with 215.23G MACs. In terms of MOTA score, TransTrack \cite{transtrack} performs slightly better for higher resolutions of \((760 \times 1266)\) and \((800 \times 1333)\), however, is computationally very heavy. For smaller resolutions, both our models outperform TransTrack \cite{transtrack} when tested on MOT17 half-half set and achieve very close MOTA score for the highest resolution. Therefore, our architecture is highly scalable across all the tested resolutions in comparison to TransTrack \cite{transtrack}.

\section{Conclusion}

We proposed a JDT MOT pipeline which is a light-weight and highly efficient version of TransTrack \cite{transtrack} architecture. It uses a transformer backbone to extract multi-scale feature maps through a feature pyramid network, which requires less resources for high input resolution than a standard CNN backbone. Our architecture solved the computational bottleneck associated with TransTrack's \cite{transtrack} encoder block, through a novel replacement of FFN network which performs channel fusion with logarithmic complexity using BT operation \cite{bft} and learns the spatial context within feature maps. The proposed architecture achieves 73.20\% MOTA on MOT17 dataset with 58.73\% less trainable parameters and 78.72\% less MACs in comparison to TransTrack \cite{transtrack}. Our method is the first work solving MOT with a fully-transformer based light-weight JDT pipeline and we expect it to provide future direction for research in domain of architecture optimization for several dense prediction tasks.

{\small
\bibliographystyle{ieee_fullname}
\bibliography{egbib}
}

\end{document}